 \documentclass[final,5p,times,twocolumn]{elsarticle}

\usepackage{amssymb}
\usepackage{lipsum}
\usepackage{tabularx}
\usepackage{float}

\newcommand{\textcite}[1]{\cite{#1}}

\makeatletter
\def\ps@pprintTitle{%
    \let\@oddfoot\@empty
    \let\@evenfoot\@empty
}
\makeatother

\begin{document}

\begin{frontmatter}



\title{Block Induced Signature Generative Adversarial Network (BISGAN): Signature Spoofing Using GANs and Their Evaluation}

\affiliation[label1]{organization={Faculty of Electrical and Computer Engineering, Chair of Fundamentals of Electrical Engineering, TUD Dresden University of Technology},
            city={Dresden},
            country={Germany}}

\affiliation[label2]{organization={School of Electrical and Computer Science (SEECS), National University of Sciences and Technology (NUST)},
            city={Islamabad},
            country={Pakistan}}

\affiliation[label3]{organization={Deep Learning Lab (DLL), National Center of Artificial Intelligence (NCAI)},
            city={Islamabad},
            country={Pakistan}}

\author[label1]{Haadia Amjad}

\author[label1]{Kilian Goeller}

\author[label1]{Steffen Seitz}

\author[label1]{Carsten Knoll}

\author[label2,label3]{Naseer Bajwa}

\author[label1]{Ronald Tetzlaff}

\author[label2,label3]{Muhammad Imran Malik}

\begin{abstract}
Deep learning is actively being used in biometrics to develop efficient identification and verification systems. Handwritten signatures are a common subset of biometric data for authentication purposes. Generative adversarial networks (GANs) learn from original and forged signatures to generate forged signatures. While most GAN techniques create a strong signature verifier, which is the discriminator, there is a need to focus more on the quality of forgeries generated by the generator model. This work focuses on creating a generator that produces forged samples that achieve a benchmark in spoofing signature verification systems. We use CycleGANs infused with Inception model-like blocks with attention heads as the generator and a variation of the SigCNN model as the base Discriminator. We train our model with a new technique that results in 80\% to 100\% success in signature spoofing. Additionally, we create a custom evaluation technique to act as a goodness measure of the generated forgeries. Our work advocates generator-focused GAN architectures for spoofing data quality that aid in a better understanding of biometric data generation and evaluation.
\end{abstract}



\begin{keyword}
Generative Adversarial Networks \sep Signature Spoofing \sep GANs Evaluation \sep Attention Mechanisms, Signature Verification



\end{keyword}

\end{frontmatter}




\section{Introduction}
\label{introduction}

Biometrics are measurements of the body and computations of human traits. Machine learning techniques commonly employ biometric authentication as a method of access control and identification. It is also used to identify people or groups who are under observation, serving as a means of surveillance. To be made use of, biometric data needs to be collectable and since it relates to human characteristics, it also needs to be unique and ever-lasting -- not subject to change. A behavioural trait utilised in automatic user verification systems within the biometric structures is the handwritten signature. Signature is taken as a non-invasive and safer option by several users since it is a common part of everyday life \textcite{1}. The unique characteristics of an individual's signature can be used for identification or verification purposes. Signature biometric data is typically captured using a digitising tablet or other electronic devices that record the pressure, speed, and trajectory of the signature. To collect handwritten signatures, researchers conduct focus groups or crowdsourcing events. This data is collected in the form of pairs, original and forged signatures \textcite{2}.

Deep learning methods can learn high-level features from raw biometric data, such as images or audio recordings. Deep learning allows extracting relevant information from biometric data without requiring manual feature engineering. For this reason, deep learning based verification systems have become widely popular offering more accuracy and robustness \textcite{3}. However, even with these strong verification systems, an attacker may be able to bypass the security check with skilled replications. One of the ways an attacker might bypass a biometric identity verification system is signature spoofing.

Signature spoofing is the deliberate creation of a deceptive signature by exploiting encryption vulnerabilities in the verification process, constituting a criminal act when employed for fraudulent purposes \textcite{4}. This involves either signing in someone else's name or tampering with a document to deceive or commit fraud. Forgeries include blind, trace-over, and skilled types. Highly accurate replicas can be detected by advanced verification algorithms, which analyse signature details for authenticity. Skilled forgeries imitating the original signature closely raise suspicion due to the absence of natural variation and typical imperfections. Verification systems use techniques like analysing stroke endpoints, intersections, infliction points, and curvature to detect such forgeries \textcite{5}.

Signature verification assesses if a person’s signature is authentic. To use signature biometrics for identification or verification purposes, the signature data is compared against previously stored signature samples. The comparison is typically done using pattern recognition algorithms that analyze the unique features of the signature, such as the shape of the letters, the spacing between the letters, and the overall rhythm and flow of the signature. Signature verification encompasses several techniques. First is the descriptive language which draws comparisons between the suspicious and a reference signature using hieroglyphic elements that represent all different kinds of signatures. Secondly, geometrical analysis is a common technique used in signature verification to compare the geometric properties of a signature with a known reference signature. This involves analysing the shape, size, position, and orientation of various signature features, such as the stroke endpoints and intersections. Thirdly, the analytical method is based on signature delineation and similarities between the components in each variation, making this approach useful in more complex scenarios \textcite{6}. This can include removing any noise or distortion from the signature image and standardizing the size and orientation of the image to prepare it for comparison. Then various features of the signature are extracted, such as the curvature. A classification algorithm determines whether a signature is genuine or forged. Apart from classification models, a generative model can also be used to distinguish between original and forged signatures by identifying underlying patterns and structures of data to reach the goal of generating similar data.

In generative modelling, the underlying distribution of a dataset is learnt, and new samples that are comparable to the original data are produced. Generative modelling is probabilistic because it involves modelling the probability distribution of the data and generating new samples from this distribution. In probabilistic modelling, the goal is to estimate the intrinsic probability distribution of the data, based on a set of observed data samples \textcite{7}. The probability distribution can then be used to generate new data samples that are similar to the observed data.

One of the most commonly used generative models is the Generative Adversarial Network (GAN) consisting of two parts: a generator and a discriminator. The generator is designed to generate new samples of the original data, while the discriminator distinguishes between the original data and the generated data, In this way, they perform adversarial roles. The generator loss measures the efficiency of the generator, penalising it for failing to fool the discriminator. Similarly, the discriminator incurs loss when it fails to differentiate between the original and false data.

In the field of signature generation, GANs have been widely explored and have shown promising results. Several papers have proposed GAN-based techniques for signature generation. These studies use GANs for data augmentation for signatures or to make stronger verification systems in the form of their discriminator. One such study by Yapıcı et al. \textcite{8} presented CycleGAN architecture for offline handwritten signature generation as a data augmentation technique. Vorugunti et al. \textcite{9} proposed OSVGAN for online signature generation employing a novel variation of Auxiliary Classifier GANs. Jiang et al. \textcite{10} introduce a stroke-aware cycle-consistent GAN architecture for signature verification. The GAN is trained to generate authentic-looking signatures while preserving the stroke-level details and characteristics.

In the research studies mentioned above, the predominant focus lies on enhancing the discriminator model's capability as a robust verifier, rather than prioritising the generation of high-quality skilled forgeries. A notable absence in these investigations is the evaluation of the forgery's quality produced by the model. These studies primarily aim to employ GANs for data augmentation in the context of signatures or to improve verification systems through discriminator model training. However, a significant aspect often overlooked is the necessity for generated forgeries to maintain a certain degree of proximity to the original sample, as forgeries should neither be excessively similar nor distinctly dissimilar. The oversight of this crucial aspect when using GANs for forgery generation underscores a notable gap in the existing research landscape. Furthermore, the absence of rigorous assessment metrics for the "quality" of generated forgeries in research dealing with generated signatures or biometric data, in general, raises questions about the intrinsic value and utility of the generated dataset itself. This observation highlights the need for dedicated research efforts to establish appropriate evaluation criteria in this domain.

Additionally, in the context of signature data, a forgery that lacks significant features of the source data does not meet the criteria for a quality forgery. Traditionally, a manual approach has involved iteratively replicating data structures until a significant similarity is achieved. Consequently, forged data encompasses a range of replicative variations of the same features. Since all forgeries seek to imitate specific data points, their aggregated variations tend to converge towards data points closely associated with the underlying biometric characteristics. This convergence presents a valuable opportunity for learning, leading to enhanced comprehension and, consequently, improved results.

Our research direction, in contrast, targets generating high-quality forgeries. In the domain of biometric data, the replicated samples must accentuate the unique characteristics of the original data. Our objective is not merely to create replicas with a measurable resemblance but rather to extract information from the original signatures that can reveal the signee's biometric traits. This automated image generation process minimises data loss and emphasises the preservation of influential data points. We introduce a generator-focused generative adversarial network that uses an Inception block concept with attention heads to produce signature forgeries that can effectively spoof a signature verification system. We train this architecture with our new paradigm-shifting training technique that focuses on adverse sample learning. Additionally, we devise an evaluation metric based on influential data points to quantify the quality of the forgery.

\section{Motivation}
For our research motivation, the following constitute research gaps:

\begin{itemize}
  \item Work on signature data using GANs has been focused on better discriminators or data augmentation. The need for generator-focused research is created to emphasise a better generation of forgeries.
  \item Since forgeries of a signature can not be too similar or dissimilar to the original sample, the generated images need a certain degree of closeness to the original image. This fact is not considered while using GANs for forgery generation and hence creates a research gap.
  \item Research work focusing on generated signatures or data, in general, does not measure the “goodness” of a forgery which questions the importance or usefulness of the generated data itself. This observation creates space for research towards appropriate evaluation metrics for this area.
\end{itemize}

\section{Literature Review}
This section comprises the literature review of various techniques approaching signature verification and generation. We have focused heavily on research works in adversarial networks, GANs, and biometric data, especially signature data.

\subsection{Adversarial Networks for Signature Generation}
Handwritten signature verification is a challenging problem in the field of biometrics and several studies have been conducted to improve its performance. To strengthen verification systems, adversarial networks have been used to generate new forgeries to adversarially attack the system. In the research work of Huan Li et al. \textcite{11}, a novel adversarial variation network (AVN) model is proposed that actively varies existing data and generates new data to mine effective features for better signature verification performance. The AVN model consists of three modules - extractor, discriminator, and variator - that work together in an adversarial way with a min-max loss function. The authors tested the proposed method on four challenging signature datasets of different languages. On CIDAR, for example, they achieve 3.77 EER.

In another paper, authors Haoyang Li et al. \textcite{12} propose a new method for attacking a handwritten signature verification system using region-restricted adversarial perturbations. The authors begin by noting that many signature verification systems are vulnerable to adversarial attacks, which can cause the system to misclassify genuine signatures as forgeries. To address this issue, the authors propose a new attack strategy that involves adding adversarial perturbations to specific regions of the signature while leaving other regions unchanged. The proposed method is designed to be a black-box attack meaning that it does not require knowledge of the inner workings of the target signature verification system.

\subsection{Generative Adversarial Networks for Signature Spoofing}
Signature spoofing aims to fail verification systems in their task of classifying genuine and forged signatures by passing high-quality skilled forgeries that get mistaken for original signatures. Some work has been done to achieve this task using GANs.

Zhang et al. \textcite{73} proposed a multi-phase system for offline signature verification using deep convolutional generative adversarial networks (DCGANs). The authors extracted local and global features from signature images using a pre-trained convolutional neural network (CNN) and used a DCGAN to generate multiple plausible variants of the signature. They combined the extracted features from the original signature image with the features extracted from the generated variants and used them for signature verification with an SVM classifier. The authors evaluated the proposed system on two publicly available signature datasets and achieved state-of-the-art performance with an equal error rate (EER) of 2.25\% and 3.06\% respectively.

Traditional methods of image recognition face challenges such as feature selection, lack of standardization, and low accuracy. A study by Wang and Jia \textcite{74} proposes a special network called SIGAN (Signature Identification GAN) based on the idea of dual learning. The trained discriminator of SIGAN is used to determine the authenticity of test handwritten signatures with the loss value of the trained discriminator serving as the identification threshold. The experimental dataset used in this study consists of five hard pen-type signatures including both genuine and deliberate imitations. The experimental results show that the average accuracy of the SIGAN-based signature identification model is 91.2\%, which is 3.6\% higher than that of traditional image classification methods.

Online Signature Verification (OSV) is an important task in the field of biometrics, which is challenging due to data scarcity and intra-writer variations. In their research work, Vorugunti et al. \textcite{9} propose a novel OSV framework that addresses these challenges using two methods. Firstly, to address the issue of data scarcity, they generate writer-specific synthetic signatures using Auxiliary Classifier GAN (AC-GAN), trained with a maximum of 40 signature samples per user. Secondly, to achieve a one-shot OSV with reduced parameters, they propose a Depth-wise Separable Convolution-based Neural Network. The authors evaluate their proposed framework on two widely used datasets, SVC and MOBISIG, and demonstrate its state-of-the-art performance in almost all categories of experimentation.

Jiajia Jiang et al. \textcite{10} presented a novel signature verification approach using a stroke-aware cycle-consistent generative adversarial network (SACGAN). This method synthesizes fake signatures with different styles and variations to augment the training data and improve the system's generalization performance. The SACGAN model is stroke-aware, meaning that it generates fake signatures with similar strokes and structures as genuine ones. Similarly, Yapıcı et al. \textcite{8} proposed a deep learning-based data augmentation method to generate synthetic signatures for improving the offline handwritten signature verification system. The proposed method uses a GAN-based data augmentation approach to create additional synthetic samples that are diverse, realistic, and representative of the signature dataset.

Since GANs have gained immense popularity in the field of computer vision for their ability to generate realistic images, Fazle Rabbi et al. \textcite{75} investigated the application of conditional GANs for generating fake images of handwritten signatures. They implemented a GAN model that can generate fake signatures by taking in a condition vector tailored by humans. Jordan Bird \textcite{76} explored how robots and generative approaches can be used for adversarial attacks on signature verification systems. They trained a convolutional neural network for signature verification and then used two robots to forge signatures to test the system's security. The results showed that the robots and conditional GAN were able to fool the system to a significant extent, but fine-tuning of the model and transfer learning with robotic and generative data reduced the attack success rate to below the model threshold.

\begin{figure}[!thb]

        \includegraphics[scale=0.40]{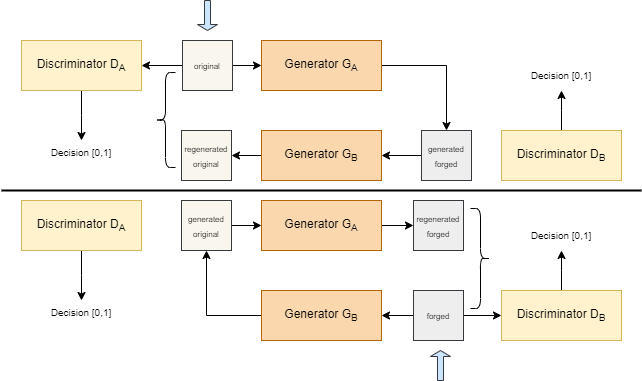}
	\caption{Overall architecture of BISGAN}
	\label{fig:fig_1}%
\end{figure}

\begin{figure*}[t]
	\centering
	\includegraphics[scale=0.50]{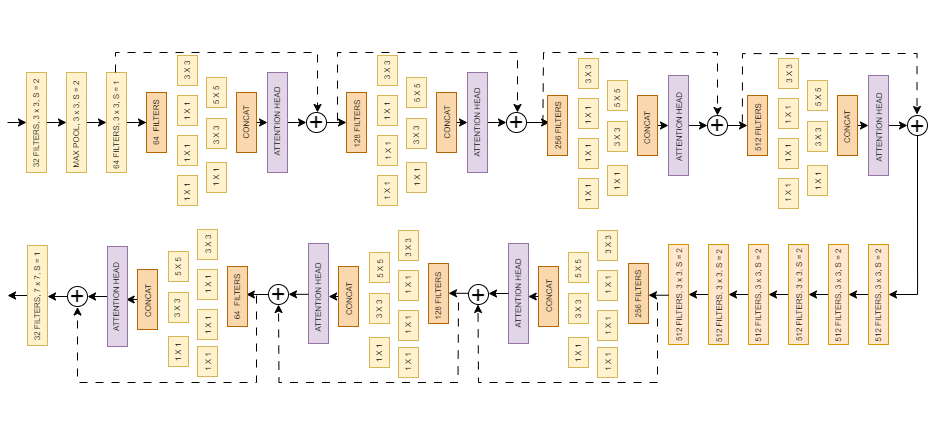}
	\caption{Generator architecture of BISGAN }
	\label{fig:gen}%
\end{figure*}

\begin{table}[!thb]
\begin{tabular}{|c|c|c|c|}
\hline\hline
\textbf{Name}                                                  & \textbf{Quantity}                                                         & \textbf{Other specs}                                                                      & \textbf{Year} \\ \hline
\begin{tabular}[c]{@{}c@{}}CEDAR \\ Signature\end{tabular}     & \begin{tabular}[c]{@{}c@{}}2640 = \\ 1320 (org) + \\ 1320 (forg)\end{tabular} & \begin{tabular}[c]{@{}c@{}}55 individuals, \\ 48 (24 + 24) \\ signatures each\end{tabular}                   & 2008          \\ \hline
\begin{tabular}[c]{@{}c@{}}SVC 2021\\ EvalDB\end{tabular} & \begin{tabular}[c]{@{}c@{}}9312 \\ = 3104 (org) \\  + 6208 \\ (forg)\end{tabular}     & \begin{tabular}[c]{@{}c@{}}75 (office) and \\  119 (mobile) \\individuals, \\ 8 (org) \\ + 16 (forg)\end{tabular}                                & 2021          \\ \hline

DeepSignDB                                                    & \begin{tabular}[c]{@{}c@{}}MCYT \\ (330 users), \\  BiosecurID \\ (400 users), \\ Biosecure DS2 \\ (650 users), \\ e-BioSign \\DS1 (65 users), e- \\ BioSign DS2 \\ (81 users)\end{tabular}   & \begin{tabular}[c]{@{}c@{}} 25 + 25 \\ 16 + 12, \\ 30 + 30 \\ 8 + 6, \\ 8 + 6 \\ (910 signatures \\ from each \\ subset)\end{tabular}                  & 2021         \\ \hline

\end{tabular}
\caption{Signature Datasets used for BISGAN training and testing}
\label{tab:signtable}
\end{table}




\section{Methodology}
To achieve high-quality generated forgeries, we create an architecture based on CycleGAN model, shown in Figure \ref{fig:fig_1}, with careful preprocessing and evaluation that suits our end goal.

\subsection{Dataset}

Many datasets are used for signature verification. These datasets contain a certain number of original signatures and a certain number of forgeries of the same user. The total images in the dataset then amount to the number of users into the sum of original and forged signatures. For our research, we have considered only English-based signatures and datasets that had no portion of synthetic images, as shown in table \ref{tab:signtable}. The usage of datasets in our work was also conditional to granted licenses.

The CEDAR Signature dataset \textcite{78} consists of 2640 signatures comprising 24 genuine signatures and 24 forged signatures for each user. It involves a total of 55 individuals with each person providing 48 signatures. The dataset was created in the year 2008 and is primarily used for handwritten signature verification tasks.

SVC2021 EvalDB \textcite{98} consists of two subsets, mobile (119 individuals) environment and office (75 individuals) environment. For all users, there are 8 genuine signatures and 16 forged signatures amounting to a total of 3104 genuine signature samples and 6208 forged samples.

DeepSignDB \textcite{99} is a combination of five (5) datasets, MCYT-300 \textcite{102}, BiosecurID \textcite{100}, Biosecure DS2 \textcite{101}, e-BioSign DS1, e-BioSign DS2. MCYT-300 consists of 25 genuine signatures and 25 forged samples for a total of 330 individuals, amounting to 16500 total images. BiosecurID contains 11200 signatures with 16 genuine samples and 12 forgeries for all 400 individuals. Biosecure DS2 contains 650 individuals with 30 genuine and 30 forged samples for all, amounting to 39000 total signatures. e-BioSign DS1 contains 8 genuine and 6 forged samples for 65 individuals amounting to 910 total images. e-BioSign DS2 also has the same 8 genuine and 6 forged ratio but for 81 users making the total number of signatures 1134. DeepSignDB contains a total of 68744 signatures. Due to computational limitations, we extract 910 images from each of the subsets maintaining the genuine to forged signature ratio of that dataset.

\subsection{Generator}
Our architecture is based on CycleGAN architecture \textcite{103}. One of the primary advantages of CycleGAN is its ability to perform unsupervised image translation, meaning it can learn to convert images from one domain to another without the need for paired training data. This flexibility makes CycleGAN particularly valuable when paired datasets are scarce or difficult to obtain. This ability of CycleGAN makes it suitable for our work. Moreover, CycleGAN can handle non-parallel data, allowing it to learn mappings between domains with distinct characteristics. The inherent cyclic consistency of CycleGAN enables the preservation of content and structure during image translation, resulting in realistic and coherent output. Each CycleGAN model consists of two generators: one for translating images from domain A to domain B and another for the reverse translation from domain B to domain A. These domains become the genuine and forged signatures in our case.

\begin{figure*}[t]
	\centering
	\includegraphics[scale= 0.50]{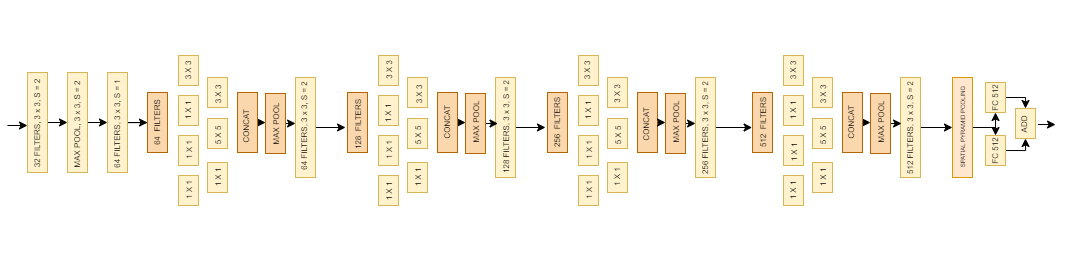}
	\caption{Discriminator of BISGAN}
	\label{fig:dis}%
\end{figure*}

CycleGAN's cyclic consistency is an efficient solution to feature learning from both domains. However, in our case, we do not wish to simply create an identical signature or one that carries some quantifiable resemblance. We want to extract the information in the original signature that may identify the biometric trait of the signee. Doing this in an automated fashion for image generation requires the least amount of data loss. It also demands influential points of the data to be preserved. To the naked eye, it seems as if a certain writing style, adding loops or combining cursive with lowercase letters, for example, is the distinct characteristic of the signature. However, these are surface-level distinctions which are also easily identified by a skilled forger. For all of the above reasons, it is important in our pipeline to include such mechanisms that increase influential data preservation.

Before using mechanisms to emphasize important features of the data, we work towards data preservation. A generator architecture consists of convolution layers before and after a transformer setup. These convolution layers are targeted for innovation to perverse data. Simple convolution layers can cause data loss due to the nature of convolution filters. Vanishing gradients can lead to slow convergence and data loss. Skip connections use regularization to resolve vanishing gradients by concatenating activations. By convention, many generators are based on ResNet or Unet. We utilize ResNet architecture for our generator with the aim of data preservation using residual blocks.

Filtering directly influences feature extraction in a neural network. To emphasize the important features of the data, it is potent to experiment with filters in a manner that forwards the best representations of data. Of course, that may call for experimentation with filter sizes \textcite{104}. A clever solution for high-level feature extraction is the use of inception blocks. Inception blocks are stacked to increase network depth, enabling the learning of hierarchical representations and capturing complex relationships within the data, leading to improved performance in various tasks. We use inception blocks because they enable multi-scale feature extraction by performing convolutions of different filter sizes in parallel, allowing the model to capture fine-grained and high-level abstract features simultaneously. After each convolution layer, we place an inception block to support them in efficient feature extraction.

Reducing data loss and enabling high-level parallel feature extraction serves our purpose. However, while the extraction of the data efficiently is guaranteed, the emphasis on the forwarded data can be increased. This emphasis is required to ensure the most influential parts of the data. This added mechanism fulfils the aim of extracting underlying biometric characteristics hidden in the signature data. Eventually, this pipeline aids the ultimate understanding required to generate quality forgeries. Attention layers allow the model to focus on the most relevant parts of the input data by assigning different attention weights to spatial locations or feature channels. Self-attention \textcite{105} in image data analysis enables the model to capture intricate spatial relationships between pixels, allowing it to focus on relevant regions and features within an image and preserve important structures. We use scaled dot-product attention, also known as self-attention, as our enhancement mechanism.

\subsection{Discriminator}

Our discriminator is inspired by the work done by Jiang et al. \textcite{10} In their work, they introduced SigCNN for signature verification using Spatial Pyramid Pooling. In a GAN architecture, the discriminator and generator tend to score against each other. The discriminator mustn't be weak in structure. For compatibility with our generator block structure, we alter SigCNN architecture with inception blocks similar to our generator architecture and use this architecture for both of the discriminators in our model. We use convolution layers of 64 filters, 128 filters, and 256 filters. Each layer is followed by an inception block with filters 1x1, 5x5, 3x3 and 3x3 max pool. Additionally, each inception block is followed by a max pool layer and a convolution layer that it had before the inception block as each layer in SigCNN is followed by a max pool and convolution layer. At the end of the model, we pass through a Spatial Pyramid Pooling layer followed by two parallel 512 fully connected layers that are then concatenated for the end result, shown in Figure \ref{fig:dis}.

\subsection{Training Paradigm Shift}
During training, CycleGAN enforces the generators to produce images that can be translated back to the original domain without significant information loss. This is implemented through the cycle consistency loss, which calculates the difference between the original input image and the image obtained by translating it to the target domain and then back to the original domain. The generators optimize this loss to ensure that the translations are consistent and coherent. Through an adversarial training process and the cycle-consistency constraint, the generators in CycleGAN learn to capture the mappings between two domains and generate high-quality images in both directions. This mechanism greatly influences CycleGAN's success in image style transformations. It is important to note that the focus is on the first domain, domain A.

\begin{figure}[!h]
	\centering
	\includegraphics[width=0.4\textwidth]{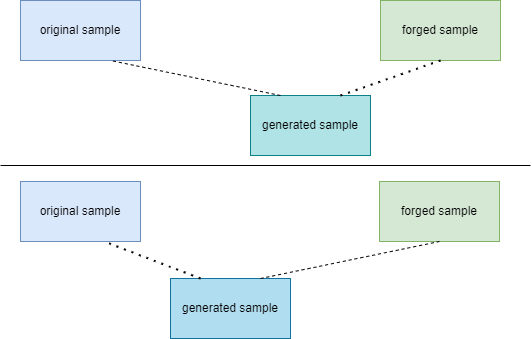}
	\caption{Abstract representation of achievement of new training technique. }
	\label{fig:train}%
\end{figure}

However, when we consider data that has deep and unique characteristics, this cycle consistency has to be altered. In merging two feature maps, where one takes precedence, we likely achieve varying outputs that may concentrate on learning features of domain B to replicate on samples of domain A during regeneration. With signature data, a generated forgery that carries fewer key features of domain A does not qualify as a quality forgery.  As discussed earlier in this work, generated forgeries can not be too similar or dissimilar to the genuine signature as a verification system would identify them as inauthentic.

For all imitated, forged and varied generated data, a manual approach has been to understand the structure of the data and replicate it continuously until a significant similarity has been achieved. Hence, forged data can be considered to contain a wide array of replications of the same features. Also, since all forgeries try to imitate certain points of the data, their variations when averaged out result in points much closer to the underlying biometric characteristic. The scope of learning from such data points to greater understanding and in turn, greater results.

When a generator learns from the latent space of an image in a domain, it learns the significant data points and aims to replicate them. If we learned from forged images instead of genuine signatures, the model would learn from the most commonly focused strong features replicated in forgeries and generate an image closer to the genuine signature, shown in Figure \ref{fig:train}.

Applying this theory to our CycleGAN-based architecture, we consider the forged images dataset domain A so that the focus is aimed in that direction. We test our theory by training our model the traditional way and also with this paradigm-shifting theory. We compare and present the results of both in the evaluation section of this paper. We see that the new training technique generates better quality forgeries than the traditional method and also achieves higher spoofing success rates.

\begin{figure*}[!thb]
	\centering
	\includegraphics[scale=0.27]{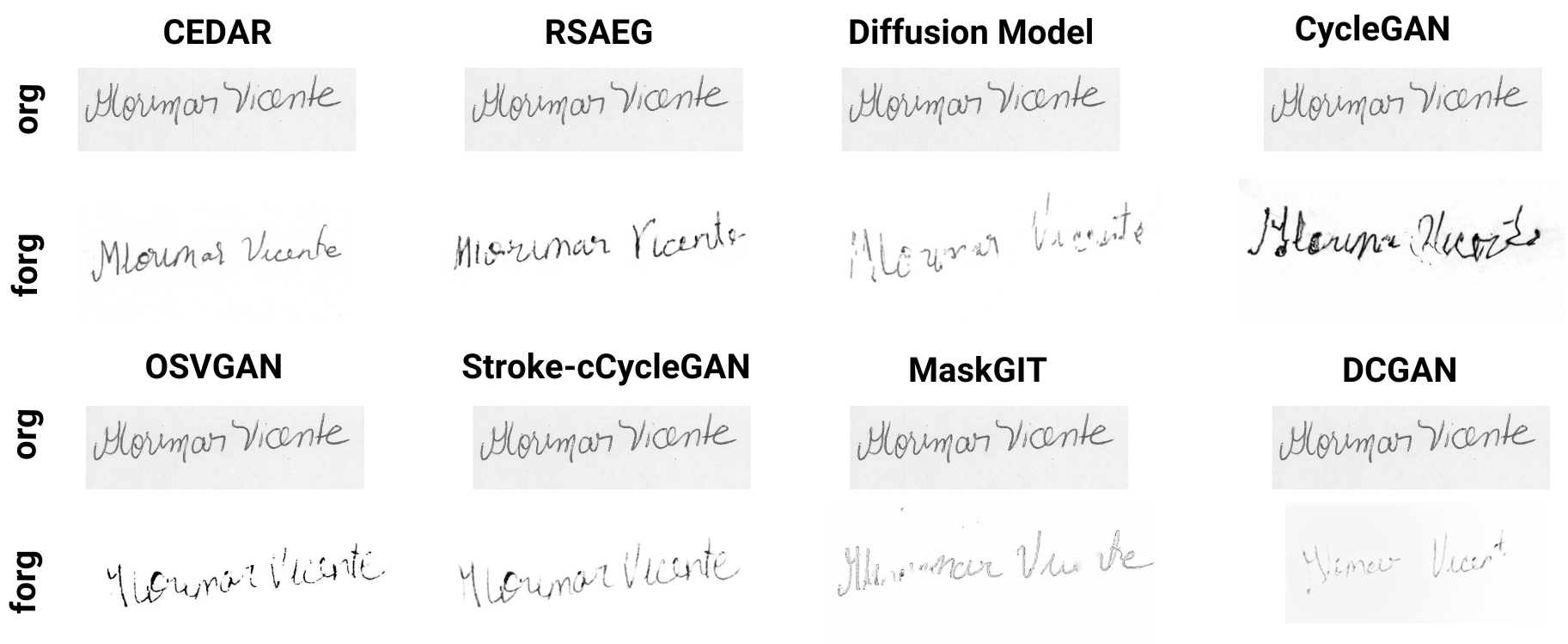}
	\caption{Comparison of generated forgeries of models.}
	\label{fig:imgsamp}%
\end{figure*}

\begin{table}[!h]
\centering
\begin{tabular}{l c c c}
 \hline
 Verification Model & ACC  & Precision \\
 \hline \hline
 CEDAR &  & \\ \hline
VGG-16 & 	0.933 & 0.917  \\
AlexNet & 0.982 & 0.947 	 \\
CapsNet & 0.887 & 0.813 	 \\
SigNet-F & 0.980 & 0.961  \\ \hline
DeepSignDB & & \\ \hline
VGG-16 & 	0.966 & 0.954  \\
AlexNet & 0.978 & 0.937 	 \\
CapsNet & 0.916 & 0.911 	 \\
SigNet-F & 0.966 & 0.952 	 \\
\hline
SVC2021\_EvalDB &  & \\ \hline
VGG-16 & 	0.944 & 0.939  \\
AlexNet & 0.992 & 0.978 	 \\
CapsNet & 0.934 & 0.921 	 \\
SigNet-F & 0.974 & 0.941 	 \\
 \hline
\end{tabular}
\caption{Performance of Verification Models on CEDAR, DeepSignDB and SVC2021\_EvalDB Signature Dataset
}
\label{tab:Table1}
\end{table}

\section{Experimentation and Evaluation}

It is important to note that the success of our work can not be completely measured with traditional metrics as the goal of our generated images is to fail the verification systems. Hence, the performance of these systems would be bad, indicating that the system is unable to correctly identify the generated forgeries as forgeries, which is the goal. We perfect other experiments to quantify the success of our model. Additionally, we propose an evaluation technique that helps present the quality of the forgery generated and can be used to define the quality of other domains of image generation.

\subsection{Spoofing Verification Systems}
Signature spoofing attempts to make a verification system unable to identify the forged signatures. As that is the goal of our model, the verification systems should perform poorly. We quantify this by analyzing the percentage of forged images that the verification system labels as genuine signatures. We brand this percentage as our success rate.

For this experiment, we train four deep learning models on CEDAR, DeepSignDB, and SVC2021 EvalDB signature datasets to act as our verification systems. It is important to note that these datasets are small for classification learning and may impact results. Regardless, we stick with these datasets because BISGAN model is trained on them. Our verification systems are VGG-16 \textcite{84}, AlexNet \textcite{85}, SigNet-F \textcite{106}, and CapsNet \textcite{86} models, shown in Table \ref{tab:Table1}. Of these three, AlexNet performs the best during traditional training and testing.

\begin{table}[!h]
\centering
\begin{tabular}{l c c c c}
 \hline
 Model & VGG-16  & AlexNet & CapsNet & SigNet-F \\
 \hline
RSAEG & 57.50\% & 57.50\% & 76.25\%  & 55\% \\
\begin{tabular}[l]{@{}l@{}}Diffusion \\ Model\end{tabular} & 28.75\% & 17.50\% &	28.75\% & 27.50\%\\
CycleGAN & 35\% & 38.75\% & 48.75\%  & 36.25\%\\
OSVGAN & 37.50\% & 37.50\% & 46.25\% & 37.50\% \\
\begin{tabular}[l]{@{}l@{}}Stroke-\\cCycleGAN\end{tabular}  & 68.75\% & 62.50\% &  76.25\% & 58.75\%\\
MaskGIT & 28.75\% & 17.50\% & 37.50\% & 27.50\% \\
DCGAN & 18.75\% & 27.50\% & 48.75\% & 18.75\%\\
\textbf{BISGAN} & \textbf{96.25\%} & \textbf{88.75\%} & \textbf{97.50\%} & \textbf{96.25\%}\\
\begin{tabular}[c]{@{}c@{}} \textbf{BISGAN (} \\ \textbf{paradigm)} \end{tabular} & \textbf{97.50\%} & \textbf{91.25\%} & \textbf{100\%} & \textbf{98.75\%}\\
 \hline
\end{tabular}
\caption{ Results of different techniques on signature verification systems, with training based on CEDAR. The percentage determines how successful the technique is in fooling the verification system. Example: if a technique has obtained 60\% success, it means that 6 out of 10 images given to the system were incorrectly identified as original signatures when in truth they were forgeries.
}
\label{tab:Table2}
\end{table}

\begin{table}[!h]
\centering
\begin{tabular}{l c c c c}
 \hline
 Model & VGG-16  & AlexNet & CapsNet & SigNet-F\\
 \hline
RSAEG & 62.50\% & 58.75\% & 58.75\% & 57.50\% \\
\begin{tabular}[l]{@{}l@{}}Diffusion \\ Model\end{tabular} & 27.50\% & 18.75\% &	27.50\% & 26.25\% \\
CycleGAN & 36.25\% & 36.25\% & 47.50\% & 36.25\%\\
OSVGAN & 47.50\% & 38.75\% & 47.50\% & 38.75\% \\
\begin{tabular}[l]{@{}l@{}}Stroke-\\cCycleGAN\end{tabular} & 67.75\% & 73.75\% &  76.25\% & 73.75\% \\
MaskGIT & 27.50\% & 18.75\% & 33.75\%  & 35\%\\
DCGAN & 17.50\% & 27.50\% & 41.25\% & 27.50\%\\
\textbf{BISGAN} & \textbf{96.25\%} & \textbf{91.25\%} & \textbf{98.75\%} & \textbf{96.25\%}  \\
\begin{tabular}[c]{@{}c@{}} \textbf{BISGAN (} \\ \textbf{paradigm)} \end{tabular} & \textbf{97.50\%} & \textbf{97.50\%} & \textbf{98.75\%} & \textbf{97.50\%} \\
 \hline
\end{tabular}
\caption{ Results of different techniques on signature verification systems, with training based on DeepSignDB.}
\label{tab:spoofotherdata}
\end{table}

\begin{table}[!h]
\centering
\begin{tabular}{l c c c c}
 \hline
 Model & VGG-16  & AlexNet & CapsNet & SigNet-F\\
 \hline
RSAEG & 61.25\% & 37.50\% & 62.50\% & 37.50\%\\
\begin{tabular}[l]{@{}l@{}}Diffusion \\ Model\end{tabular} & 27.50\% & 18.75\% &	33.75\% & 27.50\% \\
CycleGAN & 36.25\% & 36.25\% & 36.25\% & 36.25\%\\
OSVGAN & 27.50\% & 27.50\% & 36.25\% & 36.25\%\\
\begin{tabular}[l]{@{}l@{}}Stroke-\\cCycleGAN\end{tabular} & 58.75\% & 66.25\% &  77.50\% & 58.75\%\\
MaskGIT & 27.50\% & 27.50\% & 36.25\% & 27.50\%\\
DCGAN & 27.50\% & 27.50\% & 36.25\% & 36.25\%\\
\textbf{BISGAN} & \textbf{96.25\%} & \textbf{95\%} & \textbf{97.50\%} & \textbf{96.25\%}\\
\begin{tabular}[c]{@{}c@{}} \textbf{BISGAN (} \\ \textbf{paradigm)} \end{tabular} & \textbf{97.50\%} & \textbf{97.50\%} & \textbf{98.50\%} & \textbf{97.50\%} \\
 \hline
\end{tabular}
\caption{Results of different techniques on signature verification systems, with training on SVC2021\_EvalDB }
\label{tab:spooficdar}
\end{table}

\begin{figure}[!thb]
	\centering
	\includegraphics[scale=0.25]{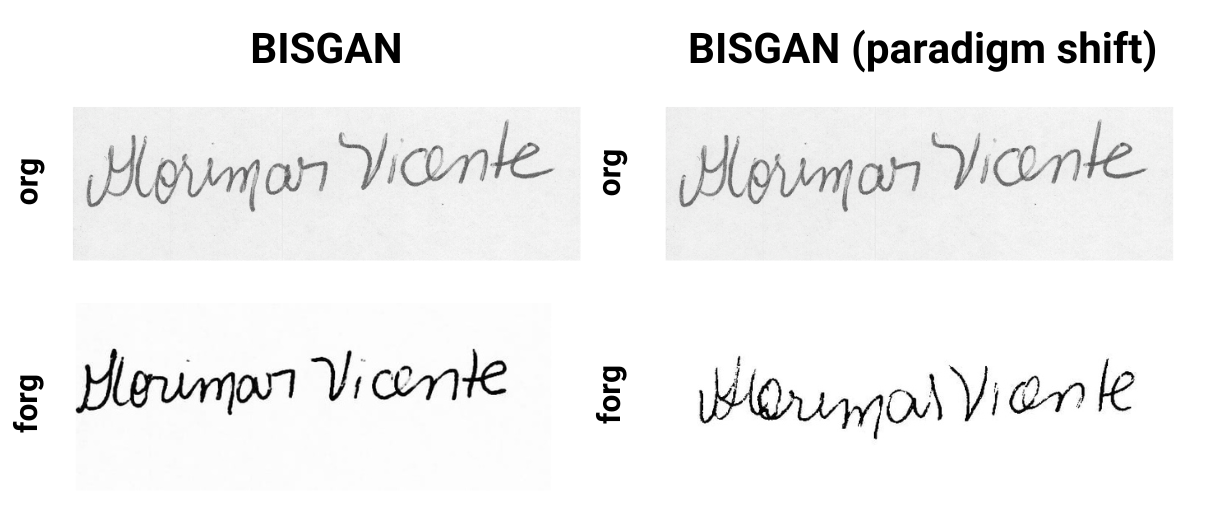}
	\caption{Forgeries generated by BISGAN compared with original samples. }
	\label{fig:bisgan}%
\end{figure}

Next, we generate ten (10) forgeries from the BISGAN model, shown in figure \ref{fig:imgsamp} and \ref{fig:bisgan}. Additionally, we train seven (7) other image generation models on CEDAR, SVC2021\_EvalDB and DeepSignDB signature datasets and generate 10 forgeries from all of these. This is also done to show the generation capabilities of BISGAN to further establish generalizability. Two (2) image generation models are based on techniques other than GANs to generate images, namely, RSAEG (perturbation-based) and the Diffusion model \textcite{87}. Two (2) of them are GAN techniques that have not been used for signature generation, namely, MaskGIT \textcite{88} and DCGAN \textcite{89}. Three (3) of them are the latest GAN techniques used to generate signatures; CycleGAN, OSVGAN and Stroke-cCycleGAN. We pass the generated images of all the above architectures one by one as input to the four (4) verification systems that we have trained. We extract the success rate of all these architectures including our own, shown in Table \ref{tab:Table2}, Table \ref{tab:spoofotherdata} and Table \ref{tab:spooficdar}.

We observe that our BISGAN with paradigm shift training performs the best towards our goal of signature spoofing followed closely by our normally trained BISGAN. The second and third successful techniques are Stroke-cCycleGAN and RSAEG respectively.

\subsection{Generated Quality Metric (GQM)}
Our work utilizes the theory that a forged signature cannot be too similar or dissimilar to a genuine signature. However, the data characteristics of a forgery should be similar to a genuine signature if it is to spoof a verification system. This answers the question of how good the generated forgery actually is. There are plenty of similarity metrics that can express the distance between genuine and forged signatures. However, the signatures' apparent similarity can be misleading regarding spoofing quality. Hence, statistical methods are typically used with data that demands a deeper mathematical comparison of data. While techniques that evaluate on data distributions of GAN inputs exist, we emphasize the influential points of the data distributions.

Generated forgeries are a result of the generative model's learning from genuine, or in our case, forged images. This learning, for GANs, starts from the latent space, which is a multi-dimensional encoding of meaningful external data representations. The external data is from the input space. The latent space speaks to the entire feature learning process and for the case of CycleGAN, there are two spaces, one for each domain. There is no doubt that the latent space for genuine signatures and forged signatures would be different. However, the latent space of the generated forgeries would be a result of what has been learned from the earlier spaces. Essentially, the latent space is a data distribution. Further extracting influential data points from the base data distribution can narrow down the core points in the data that speak to the biometric characteristics, or unique characteristics in general. In our methodology, the paradigm-shifting training technique aimed to learn underlying data features of the genuine signatures by learning from the forged images, meaning the latent space of forged images. As the spoofing success has been achieved with this technique, we conclude that the training technique has been successful in generating forgeries closer to the original samples, meaning the original latent space. To further analyze this, we develop a metric that can analyze the quality of the generated samples by measuring distances in the data distributions. We propose the Generate Quality Metric (GQM), a metric that utilizes the data distributions of the input domain and leverages influential points of the dataset to compute the closeness of the generated image which quantifies the goodness of the generated image.

Considering influential data points converts the similarity functions into a metric for goodness as it matches the important features in the data with the generated samples. GANs use the concept of latent space to learn about the input data domain. This primary concept has inspired our use of data distributions for a quality measure as well. We find the influential points over the distribution of both, the original and forged sample data using Mahalanobis distances \textcite{90}. P. C. Mahalanobis first introduced the Mahalanobis distance as the separation between a point P and a distribution D. It takes into account the covariance structure of the data to aid in locating significant deviations from the predicted distribution. Next, we compare the influential point vectors of both the original and forged samples with the influential points of the generated forged image using Cook’s distance \textcite{91}, which is the scaled change in fitted values. It measures how much removing a specific data point alters the model's estimates which is helpful as a distance measure in our case. Ultimately, we highlight which sample, original or forged, is the generated image closer to, strictly in terms of influential factors again using Mahalanobis distance.

\begin{figure}
	\centering
	\includegraphics[width=0.4\textwidth]{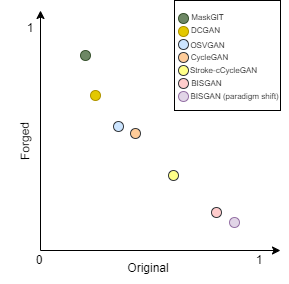}
	\caption{Mapping of GQM evaluation of generated samples of different architectures. }
	\label{fig:map}%
\end{figure}

After constructing this metric, we evaluate the generated forgeries from the GAN architectures we used in our signature spoofing experiment. We randomly pick a generated forgery from the set of ten (10) generated by each architecture. We evaluate this generated forgery using the distributions of genuine and forged samples. GQM shows the score, which is the distance value between 0 and 1, and the grade attained after comparing an image (generated forgery) to two latent spaces (generated and forged). The grade is 'O' if the sample is closer to the genuine signatures than forged signatures and 'F' if it is closer to the forged samples. The sc GQM shows BISGAN to be closest to the original, followed closely by Stroke-cCycleGAN. We map our results as shown in Figure \ref{fig:map} and Table \ref{tab:GQM}.

\begin{table}[!h]
\begin{tabular}{l c c c}
\hline\hline
\begin{tabular}[c]{@{}c@{}}GAN Models\end{tabular} & \begin{tabular}[c]{@{}c@{}}Distance \\ (genuine) \end{tabular}     & \begin{tabular}[c]{@{}c@{}}Distance \\ (forged) \end{tabular}      & GQM grade         \\ \hline\hline
\begin{tabular}[c]{@{}c@{}}CycleGAN\end{tabular}     & \begin{tabular}[c]{@{}c@{}}0.59\end{tabular} & \begin{tabular}[c]{@{}c@{}}0.41\end{tabular}                   & F        \\ \hline

\begin{tabular}[c]{@{}c@{}}OSVGAN\end{tabular} & \begin{tabular}[c]{@{}c@{}}0.52 \end{tabular}     & \begin{tabular}[c]{@{}c@{}}0.48\end{tabular}      & F          \\ \hline

Stroke-cCycleGAN   & \begin{tabular}[c]{@{}c@{}}0.37\end{tabular}   & \begin{tabular}[c]{@{}c@{}}0.63\end{tabular}                  & O          \\ \hline

MaskGIT    & \begin{tabular}[c]{@{}c@{}}0.77 \end{tabular}   & \begin{tabular}[c]{@{}c@{}}0.23\end{tabular}                  & F          \\ \hline

DCGAN    & \begin{tabular}[c]{@{}c@{}}0.69 \end{tabular}   & \begin{tabular}[c]{@{}c@{}}0.31\end{tabular}                  & F         \\ \hline

BISGAN   & \begin{tabular}[c]{@{}c@{}}0.21\end{tabular}   & \begin{tabular}[c]{@{}c@{}}0.79 \end{tabular}                  & O         \\ \hline

BISGAN (paradigm shift)   & \begin{tabular}[c]{@{}c@{}}0.12\end{tabular}   & \begin{tabular}[c]{@{}c@{}}0.88\end{tabular}                  & O          \\ \hline
\end{tabular}
\caption{GQM scores of GAN architectures trained on CEDAR.}
\label{tab:GQM}
\end{table}

\section{Discussion}
Understanding the purpose of generated images is very important to any generative AI research. In our work, understanding that signature is a biometric trait and how to replicate it to a certain threshold played an important role. We centre our work around the concept of influential points of the input data distribution while both, creating our GAN architecture and devising our evaluation metric. RSAEG proves to be an efficient technique to achieve signature spoofing. However, it is not based on GANs. Given more powerful systems to handle large amounts of data, BISGAN's training can be improved and hence its performance.\\
Our extensive evaluation techniques are proof of concept of our paradigm-shifting training technique. Including signatures generated by BISGAN into the dataset for verification systems makes them more prone to security breaches. Our research is a step in the direction of understanding the influential segments of biometric data and testing its forgeable limits. Exploiting these limitations are ethical hacking techniques to make stronger systems.
Although we believe that GQM can be generalized for many different GAN architectures since the concept of latent space is common among all, it is important to test it for different domains. For future research work, the transition of GQM to different domains and GAN architectures can be evaluated. Although we have constructed BISGAN solely for signature datasets, It could be experimented with in other domains of image translation but probably not image style transfer.

\section{Summary and Conclusions}

Signature verification encompasses various techniques, including descriptive language, geometrical analysis, and the analytical method. These methods utilize pattern recognition algorithms to compare and analyze unique features of signatures for authentication purposes. In addition to classification models, generative models are also used to differentiate between original and forged signatures by identifying underlying patterns and structures.\\
We identify a need for generator-focused research in signature data using GANs, as well as the importance of considering the percentage of similarity between original, forged, and generated samples. The lack of appropriate evaluation metrics for generated data also poses a research gap in this area.\\
Our research utilizes CycleGANs with Inception model-like blocks and attention heads, as well as the SigCNN model as a base Discriminator, to develop generators for signature forgery generation. The architecture of the generators is detailed, showcasing the combination of convolution layers, inception blocks, attention layers, and concatenation within a ResNet framework.\\
The theory that generated forgeries should possess strong features of the original signature is explored in our work and the research presents results comparing traditional training methods with a paradigm-shifting approach. We also construct a quality metric that considers the influential data points and the use of Mahalanobis distances and Cook's distance as goodness measures for generated samples. We find that the BISGAN with paradigm shift training performs the best in achieving the goal of signature spoofing, followed closely by the normally trained BISGAN.\\
To generate quality biometric data, the influential data points should be emphasized. The quality increases when adverse samples are considered for GAN training rather than genuine data samples. BISGAN's architecture covers data preservation and important feature extraction to ensure quality data generation. Biometric data generation requires domain-specific evaluation metrics that answer case-specific quality evaluation answers.

\section{Acknowledgements}
This work is partly supported by BMBF (Federal Ministry of Education and Research) in DAAD project 57616814 (https://secai.org/ SECAI, School of Embedded Composite AI) as part of the program Konrad Zuse Schools of Excellence in Artificial Intelligence.


\bibliographystyle{abbrvnat}
\bibliography{example.bib}

\begin{thebibliography}{34}
\providecommand{\natexlab}[1]{#1}
\providecommand{\url}[1]{\texttt{#1}}
\expandafter\ifx\csname urlstyle\endcsname\relax
  \providecommand{\doi}[1]{doi: #1}\else
  \providecommand{\doi}{doi: \begingroup \urlstyle{rm}\Url}\fi

\bibitem[Bird(2022)]{76}
J.~J. Bird.
\newblock Robotic and generative adversarial attacks in offline writer-independent signature verification.
\newblock \emph{arXiv preprint arXiv:2204.07246}, 2022.

\bibitem[Chang et~al.(2022)Chang, Zhang, Jiang, Liu, and Freeman]{88}
H.~Chang, H.~Zhang, L.~Jiang, C.~Liu, and W.~T. Freeman.
\newblock Maskgit: Masked generative image transformer.
\newblock In \emph{CVPR 2022}, May 2022.
\newblock URL \url{1}.

\bibitem[Dennis(1977)]{91}
C.~R. Dennis.
\newblock Detection of influential observation in linear regression.
\newblock \emph{Technometrics}, 19\penalty0 (1):\penalty0 15--18, 1977.

\bibitem[Dey et~al.(2017)Dey, Dutta, Toledo, Ghosh, Llad{\'o}s, and Pal]{106}
S.~Dey, A.~Dutta, J.~I. Toledo, S.~K. Ghosh, J.~Llad{\'o}s, and U.~Pal.
\newblock Signet: Convolutional siamese network for writer independent offline signature verification.
\newblock \emph{arXiv preprint arXiv:1707.02131}, 2017.

\bibitem[Fierrez et~al.(2010)Fierrez, Galbally, Ortega-Garcia, Freire, Alonso-Fernandez, Ramos, Toledano, Gonzalez-Rodriguez, Siguenza, Garrido-Salas, et~al.]{100}
J.~Fierrez, J.~Galbally, J.~Ortega-Garcia, M.~R. Freire, F.~Alonso-Fernandez, D.~Ramos, D.~T. Toledano, J.~Gonzalez-Rodriguez, J.~A. Siguenza, J.~Garrido-Salas, et~al.
\newblock Biosecurid: a multimodal biometric database.
\newblock \emph{Pattern Analysis and Applications}, 13:\penalty0 235--246, 2010.

\bibitem[G{\"u}nther(2014)]{4}
C.~G{\"u}nther.
\newblock A survey of spoofing and counter-measures.
\newblock \emph{NAVIGATION: Journal of the Institute of Navigation}, 61\penalty0 (3):\penalty0 159--177, 2014.

\bibitem[Hafemann et~al.(2017)Hafemann, Sabourin, and Oliveira]{2}
L.~G. Hafemann, R.~Sabourin, and L.~S. Oliveira.
\newblock Offline handwritten signature verification—literature review.
\newblock In \emph{2017 seventh international conference on image processing theory, tools and applications (IPTA)}, pages 1--8. IEEE, 2017.

\bibitem[Ho et~al.(2021)Ho, Chen, Srinivas, Li, Bachman, and Abbeel]{87}
J.~Ho, A.~Chen, A.~Srinivas, Q.~Li, P.~Bachman, and P.~Abbeel.
\newblock Denoising diffusion probabilistic models.
\newblock \emph{arXiv preprint arXiv:2006.11239}, 2021.

\bibitem[Indukuri et~al.()Indukuri, Gorthi, SriCity, and Tirupati]{9}
C.~S. V. S.~S. Indukuri, V.~P. R. K.~S. Gorthi, I.~SriCity, and I.~Tirupati.
\newblock Osvgan: Generative adversarial networks for data scarce online signature verification.

\bibitem[Jiang et~al.(2022)Jiang, Lai, Jin, Zhu, Zhang, and Chen]{10}
J.~Jiang, S.~Lai, L.~Jin, Y.~Zhu, J.~Zhang, and B.~Chen.
\newblock Forgery-free signature verification with stroke-aware cycle-consistent generative adversarial network.
\newblock \emph{Neurocomputing}, 507:\penalty0 345--357, 2022.

\bibitem[Kipouras(2022)]{5}
P.~Kipouras.
\newblock The evolution of the simulated signature by the forger.
\newblock \emph{International Journal of Law in Changing World}, 1, 2022.

\bibitem[Krizhevsky et~al.(2012)Krizhevsky, Sutskever, and Hinton]{85}
A.~Krizhevsky, I.~Sutskever, and G.~E. Hinton.
\newblock Imagenet classification with deep convolutional neural networks.
\newblock In \emph{Advances in neural information processing systems (NIPS)}, page 1097–1105, 2012.

\bibitem[Li et~al.(2021{\natexlab{a}})Li, Li, Zhang, and Yuan]{12}
H.~Li, H.~Li, H.~Zhang, and W.~Yuan.
\newblock Black-box attack against handwritten signature verification with region-restricted adversarial perturbations.
\newblock \emph{Pattern Recognition}, 111:\penalty0 107689, 2021{\natexlab{a}}.

\bibitem[Li et~al.(2021{\natexlab{b}})Li, Wei, and Hu]{11}
H.~Li, P.~Wei, and P.~Hu.
\newblock Avn: An adversarial variation network model for handwritten signature verification.
\newblock \emph{IEEE Transactions on Multimedia}, 24:\penalty0 594--608, 2021{\natexlab{b}}.

\bibitem[L{\'o}pez(2019)]{3}
A.~B. L{\'o}pez.
\newblock Deep learning in biometrics: a survey.
\newblock \emph{ADCAIJ: Advances in Distributed Computing and Artificial Intelligence Journal}, 8\penalty0 (4):\penalty0 19--32, 2019.

\bibitem[Olver et~al.(1994)Olver, Sapiro, and Tannenbaum]{6}
P.~Olver, G.~Sapiro, and A.~Tannenbaum.
\newblock \emph{Differential invariant signatures and flows in computer vision: a symmetry group approach}.
\newblock Springer, 1994.

\bibitem[Ortega-Garcia et~al.(2003)Ortega-Garcia, Fierrez-Aguilar, Simon, Gonzalez, Faundez-Zanuy, Espinosa, Satue, Hernaez, Igarza, Vivaracho, et~al.]{102}
J.~Ortega-Garcia, J.~Fierrez-Aguilar, D.~Simon, J.~Gonzalez, M.~Faundez-Zanuy, V.~Espinosa, A.~Satue, I.~Hernaez, J.-J. Igarza, C.~Vivaracho, et~al.
\newblock Mcyt baseline corpus: a bimodal biometric database.
\newblock \emph{IEE Proceedings-Vision, Image and Signal Processing}, 150\penalty0 (6):\penalty0 395--401, 2003.

\bibitem[Ortega-Garcia et~al.(2009)Ortega-Garcia, Fierrez, Alonso-Fernandez, Galbally, Freire, Gonzalez-Rodriguez, Garcia-Mateo, Alba-Castro, Gonzalez-Agulla, Otero-Muras, et~al.]{101}
J.~Ortega-Garcia, J.~Fierrez, F.~Alonso-Fernandez, J.~Galbally, M.~R. Freire, J.~Gonzalez-Rodriguez, C.~Garcia-Mateo, J.-L. Alba-Castro, E.~Gonzalez-Agulla, E.~Otero-Muras, et~al.
\newblock The multiscenario multienvironment biosecure multimodal database (bmdb).
\newblock \emph{IEEE Transactions on Pattern Analysis and Machine Intelligence}, 32\penalty0 (6):\penalty0 1097--1111, 2009.

\bibitem[Panigrahy et~al.(2009)Panigrahy, Jena, Korra, and Jena]{1}
S.~K. Panigrahy, D.~Jena, S.~B. Korra, and S.~K. Jena.
\newblock On the privacy protection of biometric traits: palmprint, face, and signature.
\newblock In \emph{Contemporary Computing: Second International Conference, IC3 2009, Noida, India, August 17-19, 2009. Proceedings 2}, pages 182--193. Springer, 2009.

\bibitem[P.C.(1936)]{90}
M.~P.C.
\newblock On the generalized distance in statistics.
\newblock \emph{Proceedings of the National Institute of Sciences of India}, 2\penalty0 (1):\penalty0 49--55, 1936.

\bibitem[Rabby et~al.(2022)Rabby, Al~Momin, and Hei]{75}
M.~F. Rabby, M.~A. Al~Momin, and X.~Hei.
\newblock Handwritten signature spoofing with conditional generative adversarial nets.
\newblock In \emph{Security, Data Analytics, and Energy-Aware Solutions in the IoT}, pages 98--110. IGI Global, 2022.

\bibitem[Radford et~al.(2015)Radford, Metz, and Chintala]{89}
A.~Radford, L.~Metz, and S.~Chintala.
\newblock Unsupervised representation learning with deep convolutional generative adversarial networks.
\newblock \emph{arXiv preprint arXiv:1511.06434}, 2015.

\bibitem[Ruthotto and Haber(2021)]{7}
L.~Ruthotto and E.~Haber.
\newblock An introduction to deep generative modeling.
\newblock \emph{GAMM-Mitteilungen}, 44\penalty0 (2):\penalty0 e202100008, 2021.

\bibitem[Sabour et~al.(2017)Sabour, Frosst, and Hinton]{86}
S.~Sabour, N.~Frosst, and G.~E. Hinton.
\newblock Dynamic routing between capsules.
\newblock In \emph{Advances in neural information processing systems (NIPS)}, page 3856–3866, 2017.

\bibitem[Simonyan and Zisserman(2014)]{84}
K.~Simonyan and A.~Zisserman.
\newblock Very deep convolutional networks for large-scale image recognition.
\newblock In \emph{Proceedings of the international conference on learning representations (ICLR)}, 2014.

\bibitem[Srinivasan et~al.(2008)Srinivasan, Srihari, and Beal]{78}
H.~Srinivasan, S.~N. Srihari, and M.~J. Beal.
\newblock Machine learning for signature verification.
\newblock In \emph{2008 19th International Conference on Pattern Recognition}, page 1–4, 2008.

\bibitem[Szegedy et~al.(2015)Szegedy, Liu, Jia, Sermanet, Reed, Anguelov, Erhan, Vanhoucke, and Rabinovich]{104}
C.~Szegedy, W.~Liu, Y.~Jia, P.~Sermanet, S.~Reed, D.~Anguelov, D.~Erhan, V.~Vanhoucke, and A.~Rabinovich.
\newblock Going deeper with convolutions.
\newblock In \emph{Proceedings of the IEEE conference on computer vision and pattern recognition}, pages 1--9, 2015.

\bibitem[Tolosana et~al.(2021{\natexlab{a}})Tolosana, Vera-Rodriguez, Fierrez, and Ortega-Garcia]{99}
R.~Tolosana, R.~Vera-Rodriguez, J.~Fierrez, and J.~Ortega-Garcia.
\newblock Deepsign: Deep on-line signature verification.
\newblock \emph{IEEE Transactions on Biometrics, Behavior, and Identity Science}, 3\penalty0 (2):\penalty0 229--239, 2021{\natexlab{a}}.

\bibitem[Tolosana et~al.(2021{\natexlab{b}})Tolosana, Vera-Rodriguez, Gonzalez-Garcia, Fierrez, Rengifo, Morales, Ortega-Garcia, Carlos Ruiz-Garcia, Romero-Tapiador, Jiang, et~al.]{98}
R.~Tolosana, R.~Vera-Rodriguez, C.~Gonzalez-Garcia, J.~Fierrez, S.~Rengifo, A.~Morales, J.~Ortega-Garcia, J.~Carlos Ruiz-Garcia, S.~Romero-Tapiador, J.~Jiang, et~al.
\newblock Icdar 2021 competition on on-line signature verification.
\newblock In \emph{Document Analysis and Recognition--ICDAR 2021: 16th International Conference, Lausanne, Switzerland, September 5--10, 2021, Proceedings, Part IV 16}, pages 723--737. Springer, 2021{\natexlab{b}}.

\bibitem[Vaswani et~al.(2017)Vaswani, Shazeer, Parmar, Uszkoreit, Jones, Gomez, Kaiser, and Polosukhin]{105}
A.~Vaswani, N.~Shazeer, N.~Parmar, J.~Uszkoreit, L.~Jones, A.~N. Gomez, {\L}.~Kaiser, and I.~Polosukhin.
\newblock Attention is all you need.
\newblock \emph{Advances in neural information processing systems}, 30, 2017.

\bibitem[Wang and Jia(2019)]{74}
S.~Wang and S.~Jia.
\newblock Signature handwriting identification based on generative adversarial networks.
\newblock In \emph{Journal of Physics: Conference Series}, number~4, page 042047. IOP Publishing, 2019.

\bibitem[Yap{\i}c{\i} et~al.(2021)Yap{\i}c{\i}, Tekerek, and Topalo{\u{g}}lu]{8}
M.~M. Yap{\i}c{\i}, A.~Tekerek, and N.~Topalo{\u{g}}lu.
\newblock Deep learning-based data augmentation method and signature verification system for offline handwritten signature.
\newblock \emph{Pattern Analysis and Applications}, 24:\penalty0 165--179, 2021.

\bibitem[Zhang et~al.(2016)Zhang, Liu, and Cui]{73}
Z.~Zhang, X.~Liu, and Y.~Cui.
\newblock Multi-phase offline signature verification system using deep convolutional generative adversarial networks.
\newblock In \emph{2016 9th international Symposium on Computational Intelligence and Design (ISCID)}, volume~2, pages 103--107. IEEE, 2016.

\bibitem[Zhu et~al.(2017)Zhu, Park, Isola, and Efros]{103}
J.-Y. Zhu, T.~Park, P.~Isola, and A.~A. Efros.
\newblock Unpaired image-to-image translation using cycle-consistent adversarial networks.
\newblock In \emph{Proceedings of the IEEE international conference on computer vision}, pages 2223--2232, 2017.

\end{thebibliography}

\end{document}